\def\buildmode{preprint}
\documentclass[11pt]{article}
\providecommand{\buildmode}{review}
\newif\ifanon
\def\reviewmode{review}
\ifx\buildmode\reviewmode \anontrue \else \anonfalse \fi
\ifanon \usepackage[review]{acl} \else \usepackage[preprint]{acl} \fi
\usepackage{times}
\usepackage{latexsym}
\usepackage[T1]{fontenc}
\usepackage[utf8]{inputenc}
\usepackage{microtype}
\usepackage{inconsolata}
\usepackage{graphicx}
\usepackage{amsmath,amssymb,booktabs,enumitem,listings,multirow,tabularx,tikz,xcolor}
\usetikzlibrary{arrows.meta,positioning,calc}
\definecolor{eligc}{HTML}{D1495B}
\definecolor{silentc}{HTML}{0E8A85}
\definecolor{mutedc}{HTML}{697386}
\definecolor{lunaplot}{HTML}{8B5CF6}
\definecolor{terraplot}{HTML}{0EA5A4}
\definecolor{solplot}{HTML}{E76F51}
\setlist{nosep,leftmargin=*}

\newcommand{\passone}{\ensuremath{\mathrm{pass@1}}}
\newcommand{\passtwo}{\ensuremath{\mathrm{pass@2}}}
\newcommand{\passsq}{\ensuremath{\mathrm{pass}^{\wedge}2}}
\newcommand{\taskcount}{87\textsuperscript{*}}
\newcommand{\artifactlink}{\ifanon an anonymized repository (link in the submission's supplementary material)\else \url{https://huggingface.co/datasets/failproofai/fire-runtime-policy-reliability}\fi}

\title{FIRE: Failure-Informed Runtime Engineering for Reliable Language-Model Agents}

\ifanon
\author{Anonymous ACL submission}
\else
\author{Nikita Agarwal \\ Failproof AI \\ \texttt{nikita@befailproof.ai} \And
        Nivedit Jain \\ Failproof AI \\ \texttt{nivedit@befailproof.ai}}
\fi

\begin{document}
\maketitle

\begin{abstract}
Language-model agents often reach a working solution and then fail to consistently deliver
it. We study runtime policies: targeted natural-language instructions and
action denials applied by the agent harness at states that preceded observed
failures, without changing model weights or the user prompt. With this,
keeping capability constant, we observe a meaningful unlock in delivered reliability.
Across the complete \taskcount-task Terminal-Bench 2.1 suite, with two attempts per task
policies increase repeated success (\passsq{}) in all three GPT-5.6 tiers:
50.6\%$\to$54.0\% for Luna, 55.2\%$\to$60.9\% for Terra, and
64.4\%$\to$73.6\% for Sol. Sol's best-of-two success changes by 1.2
points while repeated success rises by 9.2, showing that policies chiefly
convert reachable solutions into dependable delivery. We further cover 14 tasks
under Terra's frozen portfolio. Policy-guided Terra reaches 71.4\%, compared with
64.3\% for unassisted Sol, at about half the cost, demonstrating how engineering around models
could unlock dependability for a use case. To isolate the mechanism we run a randomized five-arm
experiment: real policies reach 61\% on eligible tasks, versus 39\% without a
policy, 36\% with a timing-matched sham, and 39--43\% with generic verification
or reconsideration. The intended corrective behavior appears in 22 of 24
coded policy attempts, against at most 14 in any other arm. Runtime policies
are therefore a practical reliability layer: they make capabilities an agent
already possesses substantially more repeatable.
\end{abstract}
\begingroup
\renewcommand{\thefootnote}{*}
\footnotetext{Terminal-Bench 2.1 contains 89 tasks. We uniformly exclude
\texttt{qemu-alpine-ssh} and \texttt{qemu-startup} because their pinned images
could not install the agent; every reported complete-suite comparison uses the
same remaining 87 tasks.}
\endgroup

\section{Introduction}
\label{sec:intro}

Terminal agents often fail after the hard part of a task is done. In our
policy-free runs, an agent writes a working gRPC server, confirms that it
answers, and ends its turn; the server exits with the agent's shell, and the
verifier finds nothing listening. Another agent opens the only copy of a
damaged SQLite database before copying its write-ahead log, altering the
evidence it was asked to recover. A third applies the requested replacements
exactly and then runs a formatter that changes protected bytes. In each case the
model produced the substance of the solution and the system lost it during
delivery. Such losses are common: on the \taskcount-task Terminal-Bench 2.1 population
we study, 16--19 tasks per model tier pass in exactly one of two identical
policy-free attempts (\S\ref{sec:suite}).

Two remedies are standard. One asks the model to check its own work, through
reflection, self-refinement, or more test-time computation
\citep{shinn2023reflexion,madaan2023selfrefine,snell2025scaling}. The other
enforces declared rules at runtime, as in guardrail and agent-specification
systems \citep{rebedea-etal-2023-nemo,wang2026agentspec,shi2025progent}. Both
are usually compared with an agent that receives no intervention. That
comparison cannot separate what a targeted rule says from the fact that it
interrupts the agent at a decision point: an agent stopped at the end of its
trajectory and asked to look again may repair its work whatever it is told.

Apart from two denial rules, a policy acts by placing a short natural-language
instruction in the agent's context at a chosen moment, and it decides whether
a task is in scope by reading the task description. Does the model respond to
what an in-context instruction says, or only to being interrupted? Studies of
LLM self-correction show how hard it is to separate the two
\citep{huang2024selfcorrect,kamoi-etal-2024-llms}.

We derive runtime policies from observed failures and evaluate them with a
design built to isolate their content. Each policy pairs an eligibility
predicate over the task description with a runtime predicate over the agent's
actions, and intervenes only when the trajectory reaches a state that preceded
observed failures. After freezing the portfolio we run a randomized five-arm
panel. A triggered sham fires at the same events on the same tasks but carries
generic review text. Always-on verification and reconsideration instructions
stand in for generic self-checking. Silent tasks measure off-target effects. An
attempt-level funnel records whether the risky state arose, whether the policy
fired, whether the registered behavior followed, and whether the verifier
passed.

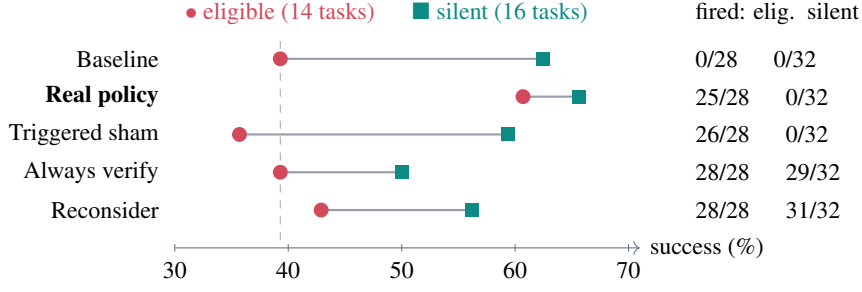
\begin{figure*}[!t]
\centering
\begin{tikzpicture}[x=0.15cm,y=0.5cm,font=\small]
  \draw[->,mutedc] (30,0) -- (71,0) node[right,black] {success (\%)};
  \foreach \x in {30,40,50,60,70} {\draw[mutedc] (\x,.1)--(\x,-.1) node[below,black] {\x};}
  \draw[mutedc!60,dashed] (39.3,.2)--(39.3,5.55);
  \foreach \name/\y/\e/\s/\fe/\fs in {
    Baseline/5/39.3/62.5/0/0,
    \textbf{Real policy}/4/60.7/65.6/25/0,
    Triggered sham/3/35.7/59.4/26/0,
    Always verify/2/39.3/50.0/28/29,
    Reconsider/1/42.9/56.2/28/31}
  {
    \node[left,align=right] at (29.5,\y) {\name};
    \draw[mutedc!70,thick] (\e,\y)--(\s,\y);
    \fill[eligc] (\e,\y) circle (2.8pt);
    \node[fill=silentc,minimum size=5pt,inner sep=0pt] at (\s,\y) {};
    \node[anchor=west] at (75,\y) {\fe/28 \quad\ \fs/32};
  }
  \node[eligc,anchor=west] at (30,6.2) {$\bullet$ eligible (14 tasks)};
  \node[silentc,anchor=west] at (50,6.2) {$\blacksquare$ silent (16 tasks)};
  \node[anchor=west,align=left] at (75,6.2) {fired: elig.\ \ silent};
\end{tikzpicture}
\caption{Randomized five-arm panel (Terra, 30 tasks, two attempts per task and
arm). Only the real portfolio raises success on the 14 eligible tasks (circles);
the triggered sham fires on as many eligible attempts but stays at baseline,
and the generic controls, which fire on nearly every attempt, do not help. The
real portfolio never fires on the 16 silent tasks (squares), whose success is
unchanged. Dashed line: baseline on eligible tasks. Contrasts with intervals
are in Table~\ref{tab:contrasts}.}
\label{fig:controls}
\end{figure*}

We argue that agent reliability is also a systems property,
not only a model property. Model weights determine which solutions are within
reach; the harness determines whether those solutions survive recurring
execution failures. A small, targeted policy layer can therefore improve
repeatability without retraining the model and, on the failure surface it
covers, recover performance otherwise associated with a more expensive tier.
Figure~\ref{fig:controls} previews the controlled result that separates policy
content from the effect of interruption.

Our contributions are:
\begin{itemize}
  \item \textbf{Attribution.} A randomized five-arm design for harness
  interventions. On eligible tasks the frozen portfolio exceeds a
  timing-matched sham by 25.0 points (95\% CI [7.1, 46.4]; $p=0.061$, the
  prespecified primary contrast) and a fresh baseline by 21.4 points
  ($p=0.032$); generic verification and reconsideration do not reproduce the
  gain (\S\ref{sec:controls}).
  \item \textbf{Mechanism.} A funnel showing that every arm meets the risky
  state equally often, but only the real policy text reliably produces the
  corrective behavior, and that correct procedure is not sufficient where a
  capability gap remains (\S\ref{sec:funnel}).
  \item \textbf{Replication.} Same-day matched replications of one policy in
  two further tiers, and complete-suite results in three tiers whose direction
  is consistent but individually imprecise (\S\ref{sec:liveness}--\ref{sec:suite}).
  \item \textbf{Reliability and model-tier leverage.} Across the full suite,
  policies increase the share of tasks solved on both attempts in every tier.
  On Terra's 14 eligible tasks, policies move the mid-tier model into the same
  performance range as unassisted Sol at about half the cost
  (\S\ref{sec:suite}--\ref{sec:tiergap}).
\end{itemize}

\section{Related Work}
\label{sec:related}

\paragraph{Agent evaluation.}
Terminal-Bench evaluates agents on containerized command-line tasks with
task-specific verifiers \citep{merrill2026terminalbenchiclr,tb21repo};
SWE-bench, AgentBench, OSWorld, and ToolSandbox cover other interactive
settings
\citep{jimenez2024swebench,liu2023agentbench,xie2024osworld,lu-etal-2025-toolsandbox}.
$\tau$-bench introduced pass$^\wedge k$, the probability of succeeding on all
$k$ trials \citep{yao2025taubench}; we report its $k{=}2$ case beside pass@$k$
\citep{chen2021evaluating}. \citet{kapoor2025agents} and HAL
\citep{kapoor2026hal} argue that agent evaluations must account for cost,
scaffold, and run-to-run variation. Our small-sample inference follows NLP
practice on significance testing and statistical power
\citep{berg-kirkpatrick-etal-2012-empirical,dror-etal-2018-hitchhikers,card-etal-2020-little}.

\paragraph{Self-correction and test-time computation.}
ReAct interleaves reasoning and acting \citep{yao2023react}; Reflexion and
Self-Refine revise behavior from verbal feedback
\citep{shinn2023reflexion,madaan2023selfrefine}; SWE-agent designs the
agent--computer interface \citep{yang2024sweagent}; AFlow searches over
workflows \citep{zhang2025aflow}; and test-time computation can substitute for
parameters \citep{snell2025scaling}. Whether models improve from generic
requests to reconsider is contested: intrinsic self-correction often fails
without external feedback
\citep{huang2024selfcorrect,kamoi-etal-2024-llms}. Our always-verify and
reconsider arms are minimal members of this family, run under the same harness.

\paragraph{Runtime enforcement and guardrails.}
The closest work is AgentSpec \citep{wang2026agentspec}, a rule language with
triggers, predicates, and enforcement for LLM agents, evaluated mainly on
preventing unsafe actions. Progent restricts tool privileges
\citep{shi2025progent}; GuardAgent and DynaGuard use a model to check actions or
outputs against policies \citep{xiang2025guardagent,hoover2026dynaguard}; NeMo
Guardrails programs conversational rails \citep{rebedea-etal-2023-nemo}; and
AgentDojo and ToolEmu evaluate risky tool use
\citep{debenedetti2024agentdojo,ruan2023toolemu}. Our policy abstraction
resembles AgentSpec's rules and classical runtime enforcement
\citep{schneider2000enforceable,bartocci2018runtime}, and the broader idea of a
simple assurance layer around an unreliable component
\citep{sha1998simplex,patterson2002roc}. We differ in where rules come from
and what evidence we require: rules are derived from observed task failures and
target delivery rather than safety, and we test whether a rule's text, rather
than its triggering, causes the improvement.

\section{Runtime Policies Derived from Failures}
\label{sec:method}

\paragraph{Reachable versus delivered success.}
\label{sec:formulation}
Task $i$ has instruction $x_i$, environment $e_i$, and verifier $V_i$. A model
$M$ in harness $H$ induces trajectories
\begin{equation}
\tau \sim \pi_{M,H}(\cdot \mid x_i,e_i),
\qquad Y_i = V_i(\tau) \in \{0,1\}.
\end{equation}
The corresponding task-level success probability is
\begin{equation}
p_i(M,H)=\Pr_{\tau\sim\pi_{M,H}}[V_i(\tau)=1].
\end{equation}
Changing $H$ can therefore change success while $M$ remains fixed. For two
attempts per task, let $S_i=Y_{i1}+Y_{i2}$. We report
\begin{align}
\passone &= \frac{1}{2N}\sum_{i=1}^{N} S_i, \\
\passtwo &= \frac{1}{N}\sum_{i=1}^{N}\mathbb{I}[S_i\geq 1], \\
\passsq  &= \frac{1}{N}\sum_{i=1}^{N}\mathbb{I}[S_i=2].
\end{align}
Here, \passone{} is mean attempt success, \passtwo{} is observed reach, and
\passsq{} is observed repeated delivery. Two attempts do not identify $p_i$.
A failed trajectory is a \emph{procedural loss} when (i) there is
evidence the agent can solve the task, such as a paired success or a working
intermediate artifact; (ii) an observable state precedes the failure; and
(iii) a bounded harness intervention could change the relevant state
transition without supplying the solution. Policies target procedural losses and
do not supply a missing algorithm or missing domain knowledge.

\paragraph{Policy abstraction.}
\label{sec:abstraction}
A runtime policy is the tuple
\begin{equation}
P=(E,G,A,R,S),
\end{equation}
where the eligibility predicate $E(x_i)$
reads the task description once. The runtime predicate $G(h_t,S_t)$ reads the
action history $h_t$ and policy state $S_t$ at harness events: prompt
submission, before each tool call, and when the agent tries to stop. The
intervention $A$ either \emph{instructs}, adding natural-language guidance and,
at stop, requiring another turn, or \emph{denies} a proposed tool call. A
release condition $R$ and a nudge limit prevent loops; tracking rules only
update $S$. Separating $E$ from $G$ separates scope from timing: broad
eligibility spends tokens and distracts successful trajectories, and a late
runtime predicate cannot prevent an irreversible action. In the frozen
portfolio, $E$ is lexical: it matches phrases in the task description.

\paragraph{Example: service persistence.}
$E$ matches task descriptions requiring a server, daemon, endpoint, or VM to
remain reachable. $S$ records launch commands, whether a launch was detached
from the agent's session, and whether a separate functional probe followed. If
the agent tries to stop after launching a server without both, $G$ holds and
the policy instructs it to relaunch in a new session with redirected streams
and probe the endpoint from a later tool call. The policy never writes the
server or chooses a protocol; it enforces the lifecycle invariant that lets the
agent's own solution survive its shell. Figure~\ref{fig:example} shows a real
instance from the panel.

\begin{figure*}[!t]
\centering
\small
\setlength{\tabcolsep}{5pt}
\begin{tabularx}{\textwidth}{@{}>{\raggedright\arraybackslash}X|>{\raggedright\arraybackslash}X@{}}
\multicolumn{1}{c|}{\textbf{Baseline attempt (fails)}} & \multicolumn{1}{c}{\textbf{Real-policy attempt (passes)}} \\
\midrule
1. writes the proto file, server, and stubs & 1. writes the proto file, server, and stubs \\
2. launches \texttt{nohup python3 server.py \&} & 2. launches \texttt{nohup python3 server.py \&}; the in-session probe fails \\
3. runs \texttt{python3 server.py} inside a tool session & 3. runs \texttt{python3 server.py} in an interactive terminal session \\
4. in-session gRPC probe: \texttt{SetVal=42; GetVal=42} & 4. in-session gRPC probe: \texttt{SetVal=42, GetVal=42} \\
5. stops: ``started the KV store gRPC server'' & 5. tries to stop; the runtime state holds and the policy instructs:
\emph{``make the evaluator-facing process survive this shell: launch it in a
new session with all standard streams redirected, then issue a separate later
command that probes the required port''} \\
& 6. \texttt{setsid -f python3 server.py </dev/null >server.log 2>\&1} \\
& 7. separate later probe: \texttt{KVStore is reachable} \\
\midrule
verifier: \texttt{test\_real\_grpc\_server\_running} fails & verifier: 7 of 7 tests pass \\
\end{tabularx}
\caption{The service-persistence chain on a panel task (\texttt{kv-store-grpc}).
Both agents run the server inside their own session and an in-session probe
succeeds. The baseline stops, and its server dies with the session. The policy
fires at the stop event, and the agent relaunches detached and probes from a
later call. Commands are abridged from the logs.}
\label{fig:example}
\end{figure*}

\paragraph{Deriving and freezing a portfolio.}
\label{sec:derivation}
We run the policy-free agent twice per task and inspect every failed attempt of
a $1/2$ task, and $0/2$ failures whose trajectory shows a concrete procedural
omission. We exclude provider and verifier faults, missing competence, and
failures without an observable pre-verifier intervention point, then cluster by
mechanism rather than task. For each cluster we register a chain
\emph{state} $X\to$ \emph{intervention} $A\to$ \emph{behavior} $B\to$
\emph{verifier condition}, write predicates that name no task, and screen
candidates on positive and nearby-negative tasks. Inert, over-broad, or harmful
candidates are revised under a new identity or retired. A family enters the
portfolio only with an observable failure state, a positive focused result in
which it fired, and an acceptable routing audit; source hashes, routing, and
analysis rules are then frozen. The frozen Terra portfolio has eight families
in seven source files and 11 rules (Table~\ref{tab:families}), preceded by 13
evaluated Terra configurations. Retired candidates include broad verify-depth,
repetition, generic scope, and quantifier policies, whose firing correlated
with breakage (Appendix~\ref{app:authoring}). Luna's frozen portfolio shares
five of these sources; Sol's was frozen earlier and uses five broader
stop-time families. No source contains task names, fixture paths, or expected
answers.

\begin{table*}[!t]
\centering
\small
\setlength{\tabcolsep}{3.5pt}
\begin{tabularx}{\textwidth}{>{\raggedright\arraybackslash}p{2.5cm}XXX}
\toprule
Family & Runtime state $X$ & Intervention $A$ & Registered behavior $B$ \\
\midrule
Service persistence & stop after launching a server without detached launch and later probe & instruct: relaunch detached, redirect streams, probe from a new call & detached launch, then separate endpoint probe \\
Deployment preservation & destructive ref, content, or reset command after a successful deployment probe & deny the command & no destructive command after the probe \\
Database evidence & mutating open or repair before a byte copy including sidecars & deny until the copy exists & complete copy precedes first mutating open \\
Exact replacement (3 rules) & write under a closed replacement map; stop after writing & instruct: byte-level allowlist; mechanical diff at stop & mechanical diff after the last write \\
LaTeX final log & stop after a build under an explicit no-overfull-box contract & instruct: search the final log for the literal warning & literal log search after the final compile \\
Binary equivalence (2 rules) & first candidate run; stop without measured comparison & instruct: compare outputs and size; change method on a miss & comparison and size measured before stop \\
Token accounting & stop on a multi-field dataset token-count request & instruct: reconcile subset, tokenizer, field subtotals & subtotals reconciled to the total \\
Enumeration & stop on an all-valid-answers request & instruct: enumerate, test, deduplicate candidates & candidate set validated before output \\
\bottomrule
\end{tabularx}
\caption{Frozen Terra policy families. Each row is a registered chain: the
runtime state the policy watches, its intervention, and the behavior marker
fixed before the control panel was run. Predicates, hashes, and source are in
Appendices~\ref{app:spec}--\ref{app:terra-policy-source}.}
\label{tab:families}
\end{table*}

\section{Evaluation Design}
\label{sec:design}

\paragraph{Setup.}
Tasks come from Terminal-Bench 2.1 with a pinned dataset digest, run in Harbor
containers \citep{tb21repo,harbor}; all task descriptions are in English. The
agent is the Codex CLI \citep[v0.146.0;][]{codexcli} at medium reasoning effort, using three
GPT-5.6 routes that we call Luna, Terra, and Sol in increasing order of success
and price. We record the provider route identifier for every attempt. We evaluate the \taskcount tasks
defined above. Timeouts and agent failures count as failures. An attempt is replaced only when it ended before a fair model
attempt, under a rule frozen in advance (Appendix~\ref{app:lineage}). Reasoning
effort, timeouts, and nudge limits are fixed across arms; nothing was tuned on
evaluation runs.

\paragraph{Four designs.}
Table~\ref{tab:designs} lists the designs in decreasing order of the causal
claim each supports. The five-arm panel uses all 14 tasks eligible for
the Terra portfolio and 16 preregistered silent tasks, five conditions, and two
attempts per task and condition, for 300 attempts. Condition order is
randomized within each replicate. The triggered sham uses the real portfolio's
eligibility predicates and event timing with generic review text of similar
length. For deny rules the sham instructs instead of denying, because repeated
generic denial can deadlock execution while keeping denial would keep the
tested mechanism; this exception was fixed before execution. Always-verify adds
a stop-time instruction to verify the evaluator-facing artifact on every task,
and reconsider adds a generic request to reconsider the solution. Behavior
markers were fixed before any panel outcome was read. The 30-task set, the
randomized schedule, and every arm's source were hashed and committed before
the first panel attempt. The silent tasks come from the development population
and are easier than the eligible ones (baseline 62.5\% vs.\ 39.3\%), so the
eligible--silent comparison mixes mechanism with headroom and we do not claim
an interaction.

\begin{table}[!ht]
\centering
\small
\setlength{\tabcolsep}{3pt}
\begin{tabularx}{\columnwidth}{@{}>{\raggedright\arraybackslash}p{1.6cm}>{\raggedright\arraybackslash}p{1.45cm}r>{\raggedright\arraybackslash}X@{}}
\toprule
Design & Tasks & Att. & Supports \\
\midrule
Five-arm panel (Terra) & 14 elig.\ + 16 silent & 300 & attribution to content vs.\ timing and generic checking \\
Matched screen (Sol, Luna) & 7 service & 84 & one mechanism across tiers, same-day baseline \\
Complete suite (all tiers) & 87 & 1{,}044 & benchmark-level reliability and cost \\
\bottomrule
\end{tabularx}
\caption{Evaluation designs, from strongest to weakest causal claim.}
\label{tab:designs}
\end{table}

\paragraph{Inference.}
Attempts on one task share prompt, environment, and verifier, so tasks are the
inferential unit. For each task we average its two attempts per condition and
compute the mean paired difference. We report 95\% percentile intervals from
10{,}000 task-level bootstrap resamples and two-sided paired sign-flip
permutation tests over tasks (100{,}000 permutations for the panel)
\citep{berg-kirkpatrick-etal-2012-empirical,dror-etal-2018-hitchhikers}. The
prespecified primary contrast is real policy minus sham on eligible tasks; all
other contrasts are secondary. Because the 14-task panel yields discrete
permutation $p$-values, we report effect sizes, confidence intervals, and
permutation tests together (Appendix~\ref{app:stats}). Analyses conditioned on policy
firing are reported only as mechanism diagnostics.

\section{Results}
\label{sec:results}

\subsection{Policy text, not timing or generic checking, drives the gain}
\label{sec:controls}

On the 14 eligible tasks the real portfolio passes 17/28 attempts (60.7\%),
against 11/28 (39.3\%) for the fresh baseline, 10/28 (35.7\%) for the
triggered sham, 11/28 (39.3\%) for always-verify, and 12/28 (42.9\%) for
reconsideration (Figure~\ref{fig:controls}, Table~\ref{tab:contrasts}). The
primary contrast, real minus sham on eligible tasks, is $+25.0$ points (95\% CI
[7.1, 46.4], $p=0.061$): a large effect with a wide interval, not a settled
one. Real minus baseline on the same tasks is $+21.4$ points ($p=0.032$).

\begin{table*}[!t]
\centering
\small
\setlength{\tabcolsep}{5pt}
\begin{tabular}{lrlrrlrlr}
\toprule
& \multicolumn{3}{c}{Eligible (14 tasks)} & \multicolumn{2}{c}{Silent (16 tasks)} & \multicolumn{3}{c}{All (30 tasks)} \\
\cmidrule(lr){2-4}\cmidrule(lr){5-6}\cmidrule(lr){7-9}
Comparator & $\Delta$ & 95\% CI & $p$ & $\Delta$ & 95\% CI & $\Delta$ & 95\% CI & $p$ \\
\midrule
Triggered sham$^\dagger$ & 25.0 & [7.1, 46.4] & .061 & 6.3 & [$-$6.3, 21.9] & 15.0 & [3.3, 28.3] & .046 \\
Baseline & 21.4 & [7.1, 35.7] & .032 & 3.1 & [$-$6.3, 12.5] & 11.7 & [3.3, 20.0] & .039 \\
Always verify & 21.4 & [7.1, 39.3] & .064 & 15.6 & [3.1, 31.3] & 18.3 & [8.3, 30.0] & .004 \\
Reconsider & 17.9 & [$-$3.6, 42.9] & .280 & 9.4 & [$-$3.1, 28.1] & 13.3 & [0.0, 26.7] & .118 \\
\bottomrule
\end{tabular}
\caption{Real policy minus each comparator, in percentage points, with 95\%
task-bootstrap intervals and paired sign-flip $p$-values. $^\dagger$Primary
contrast.}
\label{tab:contrasts}
\end{table*}

Three observations rule out the obvious alternatives. First, the sham fires on
26 of 28 eligible attempts, at the same events as the real policy, yet does no
better than baseline ($-3.6$ points, CI [$-17.9$, 10.7]): being interrupted at
the right moment does not help by itself. Second, generic self-checking applied
everywhere does not help either, consistent with the self-correction
literature \citep{huang2024selfcorrect,kamoi-etal-2024-llms}. Always-verify and
reconsider fire on 57 and 59 of 60 attempts and end at or near baseline, and
always-verify passes only 16/32 silent attempts against 20/32 for baseline.
Third, the real portfolio never fires on the 16 silent tasks and leaves their
success unchanged (21/32 vs.\ 20/32). The eligible-minus-silent interaction
($+18.8$ points, CI [$-4.9$, 42.9], $p=0.20$) is imprecise, so localization is
supporting evidence, not an established interaction. The real arm cost \$19.92
against \$23.22 for baseline and raised mean agent time from 418 to 473
seconds, almost all on eligible tasks (190 to 273 seconds).

\subsection{Same opportunity, different behavior}
\label{sec:funnel}

\begin{table}[!ht]
\centering
\setlength{\tabcolsep}{4pt}
\begin{tabular}{@{}lcccc@{}}
\toprule
Arm & State & Fired & Behavior & Pass \\
\midrule
Baseline & 25/28 & -- & 11/24 & 11/28 \\
\textbf{Real policy} & 26/28 & 25/28 & \textbf{22/24} & \textbf{17/28} \\
Triggered sham & 26/28 & 26/28 & 13/24 & 10/28 \\
Always verify & 25/28 & 28/28 & 14/24 & 11/28 \\
Reconsider & 24/28 & 28/28 & 12/24 & 12/28 \\
\bottomrule
\end{tabular}
\caption{Funnel over the 28 eligible attempts per arm: registered runtime state
observed, intervention fired, registered behavior, verifier pass. Behavior is
coded by deterministic extractors applied identically to every arm; four
attempts per arm need manual coding and are excluded from that column.}
\label{tab:funnel}
\end{table}

Table~\ref{tab:funnel} decomposes the eligible attempts. The registered runtime
state (for example, a server launched without detachment when the agent tries
to stop, or a mutating database open before any copy) arises in 24--26 of 28
attempts in every arm, so all arms face the same opportunity. What follows
differs: the registered behavior appears in 22 of 24 coded real-policy attempts
and in 11--14 of 24 elsewhere. The codes come from deterministic extractors over
the agent's command log, written before the panel ran and applied with
identical code to every arm, so they are blind to arm by construction and
involve no LLM judge. Four attempts per arm (the enumeration and
token-accounting tasks, whose markers live in output artifacts) are excluded;
even if all four were coded against the hypothesis, the contrast would be
22/28 against at most 18/28. The sham fires as often as the real policy but
produces the behavior barely more often than baseline (13 vs.\ 11). The text of
the intervention, not its occurrence, changes behavior.

Individual families show both the mechanism and its limit. Database-evidence
preservation fires, produces a complete copy before the first mutating open,
and passes on all four eligible attempts (baseline 3/4, sham 2/4). Service
persistence fires on 11 of 12 attempts, yields a detached launch followed by a
probe on 12 of 12, and passes 7/12 (baseline 4/12, sham 3/12). Binary
equivalence fires and produces the measured comparison on both attempts, yet
both fail, as in every other arm: the policy corrected the procedure but could
not supply the missing reimplementation. The behavior is necessary but not
sufficient; of the 22 real-policy attempts with the registered behavior, 12
pass.

\subsection{One policy replicates in two further tiers}
\label{sec:liveness}

To test whether the largest mechanism holds in other tiers, we compare the
service-persistence policy against a policy-free baseline on a registered seven-task screen
(five target tasks and two service tasks that already passed, three attempts per task),
separately for Sol and Luna. Both tiers ran the same source, differing only in the
nudge limit; the Terra panel uses a narrowed successor. Success rose from 7/21 to
19/21 for Sol and from 11/21 to 19/21 for Luna, with no pass-to-fail change on any task,
and the registered behavior rose from 5/21 to 20/21 and from 8/21 to 16/21. Seven
tasks allow little inference: every task that changed moved up (5 of 5 for
Sol, 4 of 4 for Luna), giving exact sign-flip $p=0.063$ and $0.125$. Without
the two subsequently excluded tasks, the counts are
4/15$\to$13/15 and 8/15$\to$14/15 (Appendix~\ref{app:liveness}).

\subsection{Complete suite: consistent direction, gains in repeated success}
\label{sec:suite}

\begin{table*}[!t]
\centering
\small
\setlength{\tabcolsep}{5pt}
\begin{tabular}{lcccrrcrr}
\toprule
Tier & \passone{} & $\Delta$ [95\% CI] & $p$ & \passtwo{} & \passsq{} & Imp./reg. & Fired & Cost \\
\midrule
Luna  & 59.8$\to$63.2 & $+$3.4 [$-$3.4, 10.3] & .43 & 69.0$\to$72.4 & 50.6$\to$54.0 & 15/10 & 12/87 & $-$5.2\% \\
Terra & 64.4$\to$70.7 & $+$6.3 [$-$1.7, 14.4] & .17 & 73.6$\to$80.5 & 55.2$\to$60.9 & 21/11 & 15/87 & $+$0.9\% \\
Sol   & 75.3$\to$80.5 & $+$5.2 [0.0, 10.9] & .09 & 86.2$\to$87.4 & 64.4$\to$73.6 & 10/4 & 81/87 & $+$47.7\% \\
\bottomrule
\end{tabular}
\caption{Complete \taskcount-task suite, two attempts per task and condition.
$\Delta$\passone{} with 95\% task-bootstrap interval and paired sign-flip $p$.
Imp./reg.: tasks with more/fewer policy successes. Fired: tasks with at least
one intervention. Tiers use different portfolios and are not pooled.}
\label{tab:suite}
\end{table*}

  The complete-suite experiment contains two attempts on every one of
  \taskcount tasks for every model and condition---1{,}044 attempts in total.
  Policies improve \passone{} in all three tiers, but their clearest effect is on
  repeated delivery. For Sol, \passsq{} rises by 9.2 points while \passtwo{}
  rises by only 1.2 points: eight more tasks pass in both attempts, while only one
  more task passes at all. The portfolio therefore converts intermittent success
  into dependable success rather than merely expanding the set of tasks solved
  once. Terra improves both reach ($+6.9$ points) and repeated success ($+5.7$
  points), while Luna moves upward on all three metrics.

  The operational evidence points in the same direction. Agent give-ups fall in
  every tier (61$\to$54, 58$\to$41, and 36$\to$25), and improved tasks
  outnumber regressed tasks in every portfolio. Cost depends on routing breadth:
  Terra's targeted portfolio fires on 15 tasks and changes total cost by only
  $+0.9$\%; Luna's fires on 12 and reduces cost by $5.2$\%; Sol's broader
  portfolio fires on 81 and increases cost by $47.7$\%.

  The complete-suite experiment establishes the cross-tier reliability pattern;
  the randomized panel isolates the effect of policy content.

\subsection{Policies bring Terra to the Sol baseline on covered tasks}
\label{sec:tiergap}

A suite-wide average dilutes a narrow policy portfolio across the many tasks
it is designed to leave untouched. The sharper comparison is therefore on the
14 tasks covered by Terra's frozen portfolio, using the same two-attempt
protocol for every condition. Unassisted Terra passes 10 of 28 attempts
(35.7\%). With policies, it passes 20 of 28 (71.4\%): the reliability layer
doubles the number of successful attempts. Unassisted Sol, the stronger and
more expensive model, passes 18 of 28 (64.3\%). On this focused cohort,
policies erase Terra's observed model-tier deficit and place it 7.1 points
above the Sol baseline while costing \$7.19 rather than \$14.48.

The estimate is necessarily imprecise at 14 tasks (Terra with policies minus
Sol baseline: 95\% task-bootstrap CI [$-$21.4, 35.7], $p=0.82$), so this is not
a claim that Terra is generally superior to Sol. It demonstrates a more useful
systems result: model upgrades and runtime policies address different sources
of failure. A stronger model expands what the agent can solve; a targeted
policy prevents known procedural failures from discarding solutions already
within reach. Where failures are recognizable and repeatable, engineering the
harness can recover an observed model-tier gap at substantially lower cost.

\section{Discussion}
\label{sec:discussion}

\paragraph{What policies change.}
Across designs, policies convert reachable solutions into repeatable delivery.
The randomized panel shows this at the attempt level, while Sol's
complete-suite pattern---\passsq{} rising by 9.2 points as \passtwo{} remains
nearly flat---shows it across the benchmark. This establishes a complementary
systems architecture: models and search produce solutions, while runtime
policies protect those solutions through execution and delivery. The leverage
is substantial. Terra with policies reaches 70.7\% across the complete suite,
closing much of the gap to the 75.3\% Sol baseline. On the 14 tasks covered by
Terra's portfolio, policy-guided Terra reaches 71.4\% versus 64.3\% for
unassisted Sol, at approximately half the cost (\S\ref{sec:tiergap}).

\section{Conclusion}
\label{sec:conclusion}

Part of what makes a capable language agent unreliable is procedural: it can
reach a solution yet fail to deliver it again. Across two attempts on the full
\taskcount-task suite, runtime policies increase repeated success in all three model
tiers, with the strongest result on Sol: \passsq{} rises from 64.4\% to 73.6\%
while \passtwo{} is nearly flat. On the tasks covered by the Terra portfolio,
the policy-guided mid-tier model reaches the same performance range as
unassisted Sol at about half the cost. The randomized controls identify the
cause: policy content changes the decisive behavior, whereas an identically
timed interruption and generic requests to verify or reconsider do not. The
power of policies is not universal intelligence. It is dependable execution:
a lightweight harness layer can preserve capabilities the model already has.

\section*{Limitations}

Every cell has two attempts per task, so \passsq{} is an observed rate, not an estimate of each task's
long-run reliability. Panel tasks come from the development population; because
the portfolio was frozen before the panel, the panel establishes a content
effect on this distribution, not generalization. All evidence comes from one
English-language benchmark, one harness (the Codex CLI), one model family, and
one reasoning effort. Models are identified by provider route rather than an
immutable checkpoint hash. We assume no material route change during the
evaluation window, but cannot independently verify that assumption. The
randomized panel supports causal attribution; cross-date complete-suite
comparisons provide supporting evidence. Portfolios are model-specific, so
tiers do not compare identical policies, and we did not run the planned
leave-one-family-out ablation. Behavior coding covers the registered marker of each family, not every
way an agent could satisfy the verifier. Authoring required expert judgement,
and researcher hours were not logged. Policies protect delivery but do not add
capability, and they can add latency, disturb prompt caching, and, when broad,
cancel their own gains.

\section*{Ethical Considerations}

This work involves no human subjects and no personal data; all tasks run in
isolated containers. Runtime policies can deny agent actions. The policies
studied here deny only destructive operations on task evidence or deployed
state, but the same mechanism could be used to restrict an agent in ways its
users do not intend, so deployed policies should be inspectable by the people
the agent acts for. Policies that make agents more persistent, for example by
keeping services running, also make agent actions more durable, which raises
the stakes of an agent acting on a mistaken goal. We release policy source,
configuration and run-selection records, per-attempt outcomes, and analysis
outputs via \artifactlink{}
so that the evaluation setup and reported results can be inspected.
\ifanon\else
\paragraph{Competing interests.} Both authors are co-founders of Failproof AI,
which develops the runtime infrastructure used to implement the policies
evaluated in this work. The authors therefore have a financial and
professional interest in the company.

\section*{Acknowledgments}
Generative AI tools were used to critique experimental designs, implement
orchestration and analysis code, support trajectory inspection, organize the
literature, generate statistical reports, and draft and edit the manuscript.
The authors specified the research questions, made the policy and evaluation
decisions, inspected generated artifacts, reran analyses from raw data, checked
every numerical claim against machine-readable outputs, and take
responsibility for the content. AI tools were not used as evaluators or as a
source of ground-truth labels.
\fi

\bibliography{custom}

\begin{thebibliography}{34}
\providecommand{\natexlab}[1]{#1}

\bibitem[{Bartocci et~al.(2018)Bartocci, Falcone, Francalanza, and
  Reger}]{bartocci2018runtime}
Ezio Bartocci, Yli{è}s Falcone, Adrian Francalanza, and Giles Reger. 2018.
\newblock \href {https://doi.org/10.1007/978-3-319-75632-5_1} {Introduction to
  runtime verification}.
\newblock In \emph{Lectures on Runtime Verification}, pages 1--33. Springer.

\bibitem[{Berg-Kirkpatrick et~al.(2012)Berg-Kirkpatrick, Burkett, and
  Klein}]{berg-kirkpatrick-etal-2012-empirical}
Taylor Berg-Kirkpatrick, David Burkett, and Dan Klein. 2012.
\newblock \href {https://aclanthology.org/D12-1091/} {An empirical
  investigation of statistical significance in {NLP}}.
\newblock In \emph{Proceedings of the 2012 Joint Conference on Empirical
  Methods in Natural Language Processing and Computational Natural Language
  Learning}, pages 995--1005, Jeju Island, Korea. Association for Computational
  Linguistics.

\bibitem[{Card et~al.(2020)Card, Henderson, Khandelwal, Jia, Mahowald, and
  Jurafsky}]{card-etal-2020-little}
Dallas Card, Peter Henderson, Urvashi Khandelwal, Robin Jia, Kyle Mahowald, and
  Dan Jurafsky. 2020.
\newblock \href {https://doi.org/10.18653/v1/2020.emnlp-main.745} {With little
  power comes great responsibility}.
\newblock In \emph{Proceedings of the 2020 Conference on Empirical Methods in
  Natural Language Processing (EMNLP)}, pages 9263--9274, Online. Association
  for Computational Linguistics.

\bibitem[{Chen et~al.(2021)}]{chen2021evaluating}
Mark Chen et~al. 2021.
\newblock Evaluating large language models trained on code.
\newblock \emph{arXiv preprint arXiv:2107.03374}.

\bibitem[{Debenedetti et~al.(2024)Debenedetti, Zhang, Balunovi{\'c},
  Beurer-Kellner, Fischer, and Tram{\`e}r}]{debenedetti2024agentdojo}
Edoardo Debenedetti, Jie Zhang, Mislav Balunovi{\'c}, Luca Beurer-Kellner, Marc
  Fischer, and Florian Tram{\`e}r. 2024.
\newblock \href {https://arxiv.org/abs/2406.13352} {{AgentDojo}: A dynamic
  environment to evaluate prompt injection attacks and defenses for {LLM}
  agents}.
\newblock In \emph{Advances in Neural Information Processing Systems}.

\bibitem[{Dror et~al.(2018)Dror, Baumer, Shlomov, and
  Reichart}]{dror-etal-2018-hitchhikers}
Rotem Dror, Gili Baumer, Segev Shlomov, and Roi Reichart. 2018.
\newblock \href {https://doi.org/10.18653/v1/P18-1128} {The hitchhiker{'}s
  guide to testing statistical significance in natural language processing}.
\newblock In \emph{Proceedings of the 56th Annual Meeting of the Association
  for Computational Linguistics (Volume 1: Long Papers)}, pages 1383--1392,
  Melbourne, Australia. Association for Computational Linguistics.

\bibitem[{{Harbor contributors}(2026)}]{harbor}
{Harbor contributors}. 2026.
\newblock Harbor: A framework for running agents in containerized environments.
\newblock \url{https://github.com/harbor-framework/harbor}.
\newblock Accessed 17 August 2026.

\bibitem[{Hoover et~al.(2026)Hoover, Baherwani, Jain, Saifullah, Vincent, Jain,
  Rad, Bruss, Panda, and Goldstein}]{hoover2026dynaguard}
Monte Hoover, Vatsal Baherwani, Neel Jain, Khalid Saifullah, Joseph Vincent,
  Chirag Jain, Melissa~Kazemi Rad, C.~Bayan Bruss, Ashwinee Panda, and Tom
  Goldstein. 2026.
\newblock \href {https://arxiv.org/abs/2509.02563} {{DynaGuard}: A dynamic
  guardian model with user-defined policies}.
\newblock In \emph{International Conference on Learning Representations}.

\bibitem[{Huang et~al.(2024)Huang, Chen, Mishra, Zheng, Yu, Song, and
  Zhou}]{huang2024selfcorrect}
Jie Huang, Xinyun Chen, Swaroop Mishra, Huaixiu~Steven Zheng, Adams~Wei Yu,
  Xinying Song, and Denny Zhou. 2024.
\newblock \href {https://openreview.net/forum?id=IkmD3fKBPQ} {Large language
  models cannot self-correct reasoning yet}.
\newblock In \emph{International Conference on Learning Representations}.

\bibitem[{Jimenez et~al.(2024)Jimenez, Yang, Wettig, Yao, Pei, Press, and
  Narasimhan}]{jimenez2024swebench}
Carlos~E. Jimenez, John Yang, Alexander Wettig, Shunyu Yao, Kexin Pei, Ofir
  Press, and Karthik Narasimhan. 2024.
\newblock {SWE-bench}: Can language models resolve real-world {GitHub} issues?
\newblock In \emph{International Conference on Learning Representations}.

\bibitem[{Kamoi et~al.(2024)Kamoi, Zhang, Zhang, Han, and
  Zhang}]{kamoi-etal-2024-llms}
Ryo Kamoi, Yusen Zhang, Nan Zhang, Jiawei Han, and Rui Zhang. 2024.
\newblock \href {https://doi.org/10.1162/tacl_a_00713} {When can {LLM}s
  actually correct their own mistakes? a critical survey of self-correction of
  {LLM}s}.
\newblock \emph{Transactions of the Association for Computational Linguistics},
  12:1417--1440.

\bibitem[{Kapoor et~al.(2025)Kapoor, Stroebl, Siegel, Nadgir, and
  Narayanan}]{kapoor2025agents}
Sayash Kapoor, Benedikt Stroebl, Zachary~S. Siegel, Nitya Nadgir, and Arvind
  Narayanan. 2025.
\newblock \href {https://arxiv.org/abs/2407.01502} {{AI} agents that matter}.
\newblock \emph{Transactions on Machine Learning Research}.

\bibitem[{Kapoor et~al.(2026)}]{kapoor2026hal}
Sayash Kapoor et~al. 2026.
\newblock \href {https://arxiv.org/abs/2510.11977} {Holistic agent leaderboard:
  The missing infrastructure for {AI} agent evaluation}.
\newblock In \emph{International Conference on Learning Representations}.

\bibitem[{Liu et~al.(2024)Liu, Yu, Zhang et~al.}]{liu2023agentbench}
Xiao Liu, Hao Yu, Hanchen Zhang, et~al. 2024.
\newblock \href {https://arxiv.org/abs/2308.03688} {{AgentBench}: Evaluating
  {LLMs} as agents}.
\newblock In \emph{International Conference on Learning Representations}.

\bibitem[{Lu et~al.(2025)Lu, Holleis, Zhang, Aumayer, Nan, Bai, Ma, Ma, Li,
  Yin, Wang, and Pang}]{lu-etal-2025-toolsandbox}
Jiarui Lu, Thomas Holleis, Yizhe Zhang, Bernhard Aumayer, Feng Nan, Haoping
  Bai, Shuang Ma, Shen Ma, Mengyu Li, Guoli Yin, Zirui Wang, and Ruoming Pang.
  2025.
\newblock \href {https://doi.org/10.18653/v1/2025.findings-naacl.65}
  {{T}ool{S}andbox: A stateful, conversational, interactive evaluation
  benchmark for {LLM} tool use capabilities}.
\newblock In \emph{Findings of the Association for Computational Linguistics:
  NAACL 2025}, pages 1160--1183, Albuquerque, New Mexico. Association for
  Computational Linguistics.

\bibitem[{Madaan et~al.(2023)Madaan, Tandon, Gupta
  et~al.}]{madaan2023selfrefine}
Aman Madaan, Niket Tandon, Prakhar Gupta, et~al. 2023.
\newblock \href {https://arxiv.org/abs/2303.17651} {Self-refine: Iterative
  refinement with self-feedback}.
\newblock In \emph{Advances in Neural Information Processing Systems}.

\bibitem[{Merrill et~al.(2026)Merrill, Shaw, Carlini, Li, Raj
  et~al.}]{merrill2026terminalbenchiclr}
Mike~A. Merrill, Alexander~G. Shaw, Nicholas Carlini, Boxuan Li, Harsh Raj,
  et~al. 2026.
\newblock \href {https://arxiv.org/abs/2601.11868} {Terminal-bench:
  Benchmarking agents on hard, realistic tasks in command line interfaces}.
\newblock In \emph{International Conference on Learning Representations}.

\bibitem[{{OpenAI}(2026)}]{codexcli}
{OpenAI}. 2026.
\newblock Codex {CLI}: Lightweight coding agent that runs in your terminal.
\newblock \url{https://github.com/openai/codex}.
\newblock Version 0.146.0; Apache-2.0 license.

\bibitem[{Patterson et~al.(2002)Patterson, Brown, Broadwell, Candea, Chen,
  Cutler, Enriquez, Fox, Kiciman, Merzbacher, Oppenheimer, Sastry, Tetzlaff,
  Traupman, and Treuhaft}]{patterson2002roc}
David Patterson, Aaron Brown, Pete Broadwell, George Candea, Mike Chen, James
  Cutler, Patricia Enriquez, Armando Fox, Emre Kiciman, Matthew Merzbacher,
  David Oppenheimer, Naveen Sastry, William Tetzlaff, Jonathan Traupman, and
  Noah Treuhaft. 2002.
\newblock Recovery-oriented computing ({ROC}): Motivation, definition,
  techniques, and case studies.
\newblock In \emph{University of California, Berkeley Technical Report
  UCB/CSD-02-1175}.

\bibitem[{Rebedea et~al.(2023)Rebedea, Dinu, Sreedhar, Parisien, and
  Cohen}]{rebedea-etal-2023-nemo}
Traian Rebedea, Razvan Dinu, Makesh~Narsimhan Sreedhar, Christopher Parisien,
  and Jonathan Cohen. 2023.
\newblock \href {https://doi.org/10.18653/v1/2023.emnlp-demo.40} {{N}e{M}o
  guardrails: A toolkit for controllable and safe {LLM} applications with
  programmable rails}.
\newblock In \emph{Proceedings of the 2023 Conference on Empirical Methods in
  Natural Language Processing: System Demonstrations}, pages 431--445,
  Singapore. Association for Computational Linguistics.

\bibitem[{Ruan et~al.(2024)Ruan, Dong, Wang, Pitis, Zhou, and
  Ba}]{ruan2023toolemu}
Yangjun Ruan, Honghua Dong, Andrew Wang, Silviu Pitis, Yongchao Zhou, and Jimmy
  Ba. 2024.
\newblock \href {https://arxiv.org/abs/2309.15817} {Identifying the risks of
  {LM} agents with an {LM}-emulated sandbox}.
\newblock In \emph{International Conference on Learning Representations}.

\bibitem[{Schneider(2000)}]{schneider2000enforceable}
Fred~B. Schneider. 2000.
\newblock \href {https://doi.org/10.1145/353323.353382} {Enforceable security
  policies}.
\newblock \emph{ACM Transactions on Information and System Security},
  3(1):30--50.

\bibitem[{Sha(2001)}]{sha1998simplex}
Lui Sha. 2001.
\newblock \href {https://doi.org/10.1109/52.936447} {Using simplicity to
  control complexity}.
\newblock \emph{IEEE Software}, 18(4):20--28.

\bibitem[{Shi et~al.(2025)Shi, He, Wang, Li, Wu, Guo, and
  Song}]{shi2025progent}
Tianneng Shi, Jingxuan He, Zhun Wang, Hongwei Li, Linyu Wu, Wenbo Guo, and Dawn
  Song. 2025.
\newblock \href {https://arxiv.org/abs/2504.11703} {{Progent}: Securing {AI}
  agents with privilege control}.
\newblock \emph{arXiv preprint arXiv:2504.11703}.

\bibitem[{Shinn et~al.(2023)Shinn, Cassano, Gopinath, Narasimhan, and
  Yao}]{shinn2023reflexion}
Noah Shinn, Federico Cassano, Ashwin Gopinath, Karthik Narasimhan, and Shunyu
  Yao. 2023.
\newblock \href {https://arxiv.org/abs/2303.11366} {Reflexion: Language agents
  with verbal reinforcement learning}.
\newblock In \emph{Advances in Neural Information Processing Systems}.

\bibitem[{Snell et~al.(2025)Snell, Lee, Xu, and Kumar}]{snell2025scaling}
Charlie Snell, Jaehoon Lee, Kelvin Xu, and Aviral Kumar. 2025.
\newblock \href {https://arxiv.org/abs/2408.03314} {Scaling {LLM} test-time
  compute optimally can be more effective than scaling model parameters}.
\newblock In \emph{International Conference on Learning Representations}.

\bibitem[{{Terminal-Bench contributors}(2026)}]{tb21repo}
{Terminal-Bench contributors}. 2026.
\newblock Terminal-bench 2.1 dataset and evaluation repository.
\newblock \url{https://github.com/harbor-framework/terminal-bench-2-1}.
\newblock Accessed 31 August 2026.

\bibitem[{Wang et~al.(2026)Wang, Poskitt, and Sun}]{wang2026agentspec}
Haoyu Wang, Christopher~M. Poskitt, and Jun Sun. 2026.
\newblock \href {https://arxiv.org/abs/2503.18666} {{AgentSpec}: Customizable
  runtime enforcement for safe and reliable {LLM} agents}.
\newblock In \emph{Proceedings of the IEEE/ACM International Conference on
  Software Engineering (ICSE)}.

\bibitem[{Xiang et~al.(2025)Xiang, Zheng, Li, Hong, Li, Xie, Zhang, Xiong, Xie,
  Yang, Song, and Li}]{xiang2025guardagent}
Zhen Xiang, Linzhi Zheng, Yanjie Li, Junyuan Hong, Qinbin Li, Han Xie, Jiawei
  Zhang, Zidi Xiong, Chulin Xie, Carl Yang, Dawn Song, and Bo~Li. 2025.
\newblock \href {https://icml.cc/virtual/2025/poster/46569} {{GuardAgent}:
  Safeguard {LLM} agents via knowledge-enabled reasoning}.
\newblock In \emph{Proceedings of the 42nd International Conference on Machine
  Learning}.

\bibitem[{Xie et~al.(2024)Xie, Zhang, Chen et~al.}]{xie2024osworld}
Tianbao Xie, Danyang Zhang, Jixuan Chen, et~al. 2024.
\newblock \href {https://arxiv.org/abs/2404.07972} {{OSWorld}: Benchmarking
  multimodal agents for open-ended tasks in real computer environments}.
\newblock In \emph{Advances in Neural Information Processing Systems}.

\bibitem[{Yang et~al.(2024)Yang, Jimenez, Wettig et~al.}]{yang2024sweagent}
John Yang, Carlos~E. Jimenez, Alexander Wettig, et~al. 2024.
\newblock \href {https://arxiv.org/abs/2405.15793} {{SWE-agent}: Agent-computer
  interfaces enable automated software engineering}.
\newblock In \emph{Advances in Neural Information Processing Systems}.

\bibitem[{Yao et~al.(2025)Yao, Shinn, Razavi, and Narasimhan}]{yao2025taubench}
Shunyu Yao, Noah Shinn, Pedram Razavi, and Karthik Narasimhan. 2025.
\newblock \href
  {https://proceedings.iclr.cc/paper_files/paper/2025/hash/1b126cc38b8638e07bef37e7b2bb72bf-Abstract-Conference.html}
  {$\tau$-bench: A benchmark for tool-agent-user interaction in real-world
  domains}.
\newblock In \emph{International Conference on Learning Representations}.

\bibitem[{Yao et~al.(2023)Yao, Zhao, Yu et~al.}]{yao2023react}
Shunyu Yao, Jeffrey Zhao, Dian Yu, et~al. 2023.
\newblock \href {https://arxiv.org/abs/2210.03629} {{ReAct}: Synergizing
  reasoning and acting in language models}.
\newblock In \emph{International Conference on Learning Representations}.

\bibitem[{Zhang et~al.(2025)Zhang, Xiang, Yu, Teng, Chen, Chen, Zhuge, Cheng,
  Hong, Wang, Zheng, Liu, Luo, and Wu}]{zhang2025aflow}
Jiayi Zhang, Jinyu Xiang, Zhaoyang Yu, Fengwei Teng, Xionghui Chen, Jiaqi Chen,
  Mingchen Zhuge, Xin Cheng, Sirui Hong, Jinlin Wang, Bingnan Zheng, Bang Liu,
  Yuyu Luo, and Chenglin Wu. 2025.
\newblock \href {https://openreview.net/forum?id=z5uVAKwmjf} {{AFlow}:
  Automating agentic workflow generation}.
\newblock In \emph{International Conference on Learning Representations}.

\end{thebibliography}

\appendix

\section{Reproducibility}
\label{app:repro}
The artifact (\artifactlink{}) contains configuration and run-selection records
with dataset, model-route, harness, and policy details; task-content digests;
per-attempt verifier outcomes, policy
firing indicators, and registered behavior labels; the frozen policy source
and hashes (Appendix~\ref{app:spec}); the selected-attempt ledger, replacement
lineage, and deviations (Appendix~\ref{app:lineage}); and machine-readable
analysis outputs. These records support recomputation of the reported
aggregate results without querying the model again. Running the policies on a
successor route is a new evaluation cell, not a replication.
Terminal-Bench 2.1, Harbor, and the Codex CLI are used under their Apache-2.0
licenses and for their intended purpose, agent evaluation. We release our
policy source and analysis code under the Apache License 2.0. We release our
original manifests, annotations, derived
data, and per-attempt outcome records under CC BY 4.0. Third-party content
remains subject to its original license.

\section{Statistical Procedure}
\label{app:stats}
For each task and condition we take the mean of its two attempts. An effect is
the mean paired task-level difference. Percentile intervals use 10{,}000
task-level bootstrap resamples. Permutation tests flip the sign of each
task-level difference under the sharp null, with 100{,}000 repetitions for the
panel and exact enumeration for the seven-task screens. The
eligible-minus-silent interaction resamples tasks within cohorts. Attempts from
one task are never treated as independent units. The bootstrap interval and
the sign-flip test answer different questions and can disagree in small
discrete samples; in the 14-task primary contrast the interval excludes zero
while $p=0.061$. We report both rather than selecting the more favorable one.

\section{Control Panel Lineage and Deviations}
\label{app:lineage}
Provider rate limits caused no-fair-attempt failures in the second replicate.
Before recovery, the replacement rule was frozen: keep each original fair
observation, and otherwise take the first fair result from the chronological
recovery chain. Replicate~1 is unchanged in every arm. For replicate~2 this
selects 25 original, 3 first-recovery, and 2 second-recovery observations for
baseline; 29 first-recovery and 1 second-recovery for the real policy; 30
first-recovery for the sham; 29 first-recovery and 1 second-recovery for
always-verify; and 30 second-recovery for reconsider. The machine-readable
lineage audit lists every candidate and decision. The two tasks identified by
the complete-suite asterisk are excluded uniformly from every cell. A repeated provider stream
failure on one task is counted as a failure under the declared rule rather than
retried selectively.

\section{Complete-Suite Details}
\label{app:suite}

\begin{table}[!htbp]
\centering
\small
\setlength{\tabcolsep}{2.6pt}
\begin{tabular}{@{}l|ccc|ccc|ccc@{}}
\toprule
& \multicolumn{3}{c|}{Luna} & \multicolumn{3}{c|}{Terra} & \multicolumn{3}{c}{Sol} \\
Base & 0 & 1 & 2 & 0 & 1 & 2 & 0 & 1 & 2 \\
\midrule
0/2 & 18 & 6 & 3 & 11 & 8 & 4 & 8 & 1 & 3 \\
1/2 & 4 & 6 & 6 & 3 & 4 & 9 & 3 & 10 & 6 \\
2/2 & 2 & 4 & 38 & 3 & 5 & 40 & 0 & 1 & 55 \\
\bottomrule
\end{tabular}
\caption{Task transitions between observed states (rows: baseline successes of
2; columns: policy successes of 2).}
\label{tab:transitions}
\end{table}

\begin{table}[!htbp]
\centering
\small
\setlength{\tabcolsep}{2.5pt}
\begin{tabular}{@{}llrrrrrr@{}}
\toprule
Tier & Cond. & Pass & Gave up & Fail & T/O & Crash & \$ / s \\
\midrule
Luna & base & 104 & 61 & 8 & 1 & 0 & 5.20 / 321 \\
     & policy & 110 & 54 & 6 & 4 & 0 & 4.93 / 382 \\
Terra & base & 112 & 58 & 2 & 1 & 1 & 49.57 / 176 \\
      & policy & 123 & 41 & 6 & 2 & 2 & 49.99 / 355 \\
Sol & base & 131 & 36 & 2 & 3 & 3 & 111.38 / 289 \\
    & policy & 140 & 25 & 3 & 4 & 4 & 164.50 / 404 \\
\bottomrule
\end{tabular}
\caption{Terminations, total cost (\$), and mean agent time (s) over 174
attempts per cell.}
\label{tab:terminations}
\end{table}

\paragraph{Transition structure.}
Of Sol's 19 baseline $1/2$ tasks, six become $2/2$ and three become
$0/2$, and 55 of its 56 baseline $2/2$ tasks stay there.

\begin{figure}[!htbp]
\centering
\resizebox{\columnwidth}{!}{\input{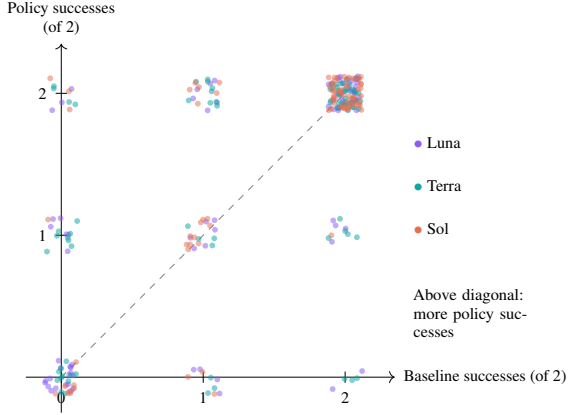}}
\caption{Task-level movement between observed states, jittered. Points above
the diagonal gain policy successes.}
\label{fig:scatter}
\end{figure}

\begin{table}[!htbp]
\centering
\small
\begin{tabular}{@{}lrrr@{}}
\toprule
Configuration & Passes & Success & Cost \\
\midrule
Terra, no policy & 10/28 & 35.7\% & \$5.85 \\
Terra $+$ portfolio & 20/28 & 71.4\% & \$7.19 \\
Sol, no policy & 18/28 & 64.3\% & \$14.48 \\
Sol $+$ portfolio & 24/28 & 85.7\% & \$27.17 \\
\bottomrule
\end{tabular}
\caption{The 14 tasks eligible for the Terra portfolio, from the complete-suite
runs (two attempts per task; each tier with its own portfolio). Terra with its
portfolio minus policy-free Sol: $+$7.1 points, 95\% CI [$-$21.4, 35.7],
$p=0.82$.}
\label{tab:tiers}
\end{table}

\section{Service-Persistence Screens}
\label{app:liveness}
\begin{table}[!htbp]
\centering
\small
\setlength{\tabcolsep}{3pt}
\begin{tabular}{@{}lrrrr@{}}
\toprule
& \multicolumn{2}{c}{Sol} & \multicolumn{2}{c}{Luna} \\
\cmidrule(lr){2-3}\cmidrule(lr){4-5}
Task & base & policy & base & policy \\
\midrule
configure-git-webserver & 0/3 & 1/3 & 0/3 & 3/3 \\
hf-model-inference & 1/3 & 3/3 & 3/3 & 3/3 \\
kv-store-grpc & 0/3 & 3/3 & 2/3 & 3/3 \\
pypi-server & 0/3 & 3/3 & 0/3 & 2/3 \\
qemu-alpine-ssh & 0/3 & 3/3 & 0/3 & 2/3 \\
nginx-request-logging & 3/3 & 3/3 & 3/3 & 3/3 \\
qemu-startup & 3/3 & 3/3 & 3/3 & 3/3 \\
\midrule
Total & 7/21 & 19/21 & 11/21 & 19/21 \\
\bottomrule
\end{tabular}
\caption{Matched service-persistence screens (three attempts per task; same-day
policy-free baseline). The last two tasks are controls that already passed.}
\label{tab:screens}
\end{table}
All 84 attempts completed without infrastructure exceptions. Every policy
attempt carried the expected model and profile marker, and no baseline attempt
contained a policy hook directory. Sol's give-ups fell from 14 to 2. The
remaining Sol failures are deploy-hook logic failures with the service
reachable, which are outside the policy's mechanism.

\section{Policy Authoring and Retirement}
\label{app:authoring}
The repository records 13 Terra policy-arm configurations with scored
artifacts, which include early bundles, isolated screens, and revisions. Broad
quantifier, repetition, generic scope, generic verification, and optimization
prompts were retired or replaced after inert or adverse results. Exact
replacement, database evidence, and final-log validation were narrowed after
false routing or late firing: for example, a phrase meaning ``one per line''
alone falsely routed a chess task into exactness checks, and the database guard
initially missed Python SQLite opens. Failures in large-scale editing, video
processing, and query optimization were classified as capability gaps.
Policies for chess legality, spectral fitting, G-code decoding, primer design,
semantic retrieval, and distributed numerical equivalence were not written
because no reliable harness-observable intervention point was found.
Luna's frozen portfolio shares five Terra sources (service, database evidence,
replacements, LaTeX, and a source-build provenance check). Sol's uses five
broader stop-time families (service, measured acceptance criteria,
public-invocation contract, requirements, scope). Researcher hours were not
logged contemporaneously; commit timestamps give chronology, not labor.

\section{Frozen Terra Policy Specification}
\label{app:spec}
Policies run in a hook layer around Codex and observe prompt-submission,
pre-tool-use, and stop events. Per-session state is isolated per task
container. The Terra profile allows one nudge per named rule, except that the
database-evidence denial repeats until a sufficient copy exists. When several
families are eligible, all tracking rules run and the hook log records every
decision, so co-firing is kept rather than assigned post hoc. The runner
refuses to score a policy attempt with no pre-tool-use hook events, a policy
load error, or a missing model and profile marker. Prompt routing was audited
over all development tasks before the final runs.

\begin{table}[!htbp]
\centering
\small
\begin{tabular}{@{}ll@{}}
\toprule
Artifact & SHA-256 (prefix) \\
\midrule
Arm configuration & \texttt{865a132e3f07} \\
Policy profiles & \texttt{62132159f374} \\
Service/deployment source & \texttt{fae2fd0c5b4f} \\
Database-evidence source & \texttt{ef67adb180fc} \\
Exact-replacement source & \texttt{4b9f61d57a04} \\
LaTeX-validation source & \texttt{0aa683406b49} \\
Binary-equivalence source & \texttt{db861cfb21c1} \\
Token-accounting source & \texttt{95eb9971c210} \\
Enumeration source & \texttt{4b6bd757abb8} \\
\bottomrule
\end{tabular}
\caption{SHA-256 prefixes of the frozen Terra treatment.}
\label{tab:hashes}
\end{table}

\section{Complete Terra Runtime-Policy Source}
\label{app:terra-policy-source}
The following files are included verbatim\ifanon, with the hook library's
package name redacted for anonymous review\fi. Their hashes are in
Table~\ref{tab:hashes}.
\lstset{basicstyle=\ttfamily\footnotesize,breaklines=true,columns=fullflexible}

\subsection{Service persistence and deployment preservation}
\begin{lstlisting}
// Luna v4: durable service lifecycle, with both prompt and command evidence.
//
// v3's Stop hook fired on 82/178 trials because any Python or Node script was
// treated as a service launcher. This version requires an explicit persistent
// service contract in the prompt AND a recognizable server/VM launch command.

import { customPolicies, allow, deny, instruct } from "failproofai";
import { load, save, commandOf, promptOf, shouldNudge } from "../_lib/state.mjs";

const PERSISTENT_SERVICE =
  /\b(?:web\s*server|server|service|daemon|listener|endpoint|ssh|telnet|socket|virtual machine|vm)\b[^.\n]{0,180}\b(?:run|running|listen|listening|reachable|available|accessible|connect|background|port|should be able)\b|\b(?:leave|keep|remain|stay)\b[^.\n]{0,80}\b(?:running|listening|reachable|available)\b|\bshould be able to (?:run\s+)?["'`]*(?:ssh|curl|telnet|connect)\b/i;

// Do not classify arbitrary `python foo.py` or `node foo.js` executions as
// services. The command must name a conventional server/VM launcher or contain
// an explicit listen/serve flag.
const SERVICE_LAUNCH =
  /\b(?:qemu-system-[\w-]+|uvicorn|gunicorn|flask\s+run|python(?:3)?\s+-m\s+http\.server|nginx|redis-server|grpc[_-]?(?:server|serve)|npm\s+(?:run\s+)?start)\b|\b(?:python(?:3)?|node)\b[^;&|\n]{0,160}\S*(?:server|serve|service|daemon|listen)[\w.-]*\.(?:py|js)\b|--(?:host|port|listen)\b/i;

const DETACHED = /\bsetsid\b|\bsystemd-run\b|\bscreen\s+-d|\btmux\s+new-session\s+-d/;
const PROBE =
  /\b(?:curl|wget|grpcurl|ssh|sshpass)\b|\bnc\s+-z|\bss\s+-l|\blsof\s+-i|socket\.create_connection|channel_ready_future|requests\.(?:get|post|put|delete)/i;
const DEPLOYMENT_TASK =
  /\b(?:bare\s+git|git\s+repository|git\s+server)\b[\s\S]{0,500}(?:\b(?:web\s*server|post-receive)\b|\bport\s+8080\b|curl\s+https?:\/\/)/i;
const DEPLOYMENT_PROBE = /\bgit\s+push\b|\bcurl\b[^\n]{0,200}\bhello\.html\b/i;
const DESTROY_DEPLOYMENT =
  /\b(?:git\s+--git-dir=\S+\s+)?update-ref\s+-d\b|\brm\s+(?:-[A-Za-z]+\s+)*[^\n;&|]*(?:hello\.html|\/var\/www\/|\/git\/server)|\bgit\s+(?:reset\s+--hard|clean\b)/i;

customPolicies.add({
  name: "luna4-service-track",
  match: { events: ["UserPromptSubmit", "PreToolUse"] },
  fn: async (ctx) => {
    const s = load(ctx);
    const prompt = promptOf(ctx);
    if (prompt && PERSISTENT_SERVICE.test(prompt)) s.l4PersistentService = true;
    if (prompt && DEPLOYMENT_TASK.test(prompt)) s.l4DeploymentTask = true;

    const cmd = commandOf(ctx);
    if (cmd) {
      s.l4Seq = (s.l4Seq || 0) + 1;
      if (s.l4PersistentService && SERVICE_LAUNCH.test(cmd)) {
        s.l4Launch = s.l4Seq;
        if (DETACHED.test(cmd)) s.l4Detached = s.l4Seq;
      }
      if (s.l4Launch && PROBE.test(cmd) && s.l4Seq > s.l4Launch) {
        s.l4ProbeAfter = s.l4Seq;
      }
      if (s.l4DeploymentTask && DEPLOYMENT_PROBE.test(cmd)) s.l4DeploymentProbe = s.l4Seq;
    }
    save(ctx, s);
    return allow();
  },
});

customPolicies.add({
  name: "luna4-preserve-verified-deployment",
  match: { events: ["PreToolUse"] },
  fn: async (ctx) => {
    const s = load(ctx);
    const cmd = commandOf(ctx);
    if (!s.l4DeploymentTask || !s.l4DeploymentProbe || !cmd || !DESTROY_DEPLOYMENT.test(cmd)) return allow();
    return deny(
      `Blocked cleanup of the deployment state after an evaluator-facing push or HTTP probe. Leave the tested ` +
      `branch, commit, deployed file, document root, and running server intact; the verifier consumes that final ` +
      `state. Narrow the command only if its target is unambiguously unrelated temporary data.`
    );
  },
});

customPolicies.add({
  name: "luna4-service-persistence",
  match: { events: ["Stop"] },
  fn: async (ctx) => {
    const s = load(ctx);
    if (!s.l4PersistentService || !s.l4Launch) return allow();
    if (s.l4Detached && s.l4ProbeAfter) return allow();
    if (!shouldNudge(s, "luna4-service", 1)) return allow();
    save(ctx, s);
    return instruct(
      `The task explicitly requires a persistent service or VM, and you launched one. ` +
      `Before finishing, make the evaluator-facing process survive this shell: launch it in a new ` +
      `session with all standard streams redirected, then issue a separate later command that probes ` +
      `the required port or endpoint. A live PID or a probe chained into the launch command is not ` +
      `enough. If the detached process already passes that later probe, stop without changing it.`
    );
  },
});
\end{lstlisting}
\subsection{Database evidence preservation}
\begin{lstlisting}
// Luna v4: prevent destruction of damaged-state evidence before inspection.
//
// This preserves the narrow, high-consequence half of v3's evidence policy.
// The broad prompt-time enumeration instruction is intentionally removed: it
// fired on both chess regressions despite not identifying their real failure.

import { customPolicies, allow, deny } from "failproofai";
import { load, save, commandOf, promptOf } from "../_lib/state.mjs";

const RECOVERY_TASK =
  /\b(?:corrupt\w*|damaged|encrypted|recover\w*|repair\w*|forensic\w*|salvage|truncated|deleted|lost)\b/i;
const BACKUP = /\b(?:cp|rsync|tar|dd|cpio|zip)\b/;
const DATABASE_TASK = /\b(?:database|sqlite|wal)\b/i;
// A database backup is complete only if the command preserves the surrounding
// directory or explicitly includes sidecars. Copying just `main.db` is not
// enough: opening it can replay and delete the only useful `main.db-wal`.
const DATABASE_BACKUP =
  /(?:\b(?:cp|rsync)\b[^\n]*(?:\/app(?:\/\.?)?\s|\.db(?:\*|\{[^}]*-wal)|\.db-wal)|\b(?:tar|zip|cpio)\b[^\n]*\/app\b)/i;
const MUTATING_OPEN =
  /(?:\bsqlite3\s+\S+\.(?:db|sqlite3?)|\b(?:python3?|pypy3?)\b[\s\S]{0,1000}?(?:import\s+sqlite3|sqlite3\.connect)|\bmysql\b|\bpsql\b|\bmongo\b|\bredis-cli\b|\bunzip\s+-o|\bmount\b|\bfsck\b|\bgit\s+(?:checkout|reset|clean|gc)|\btruncate\b)/i;

customPolicies.add({
  name: "luna4-preserve-track",
  match: { events: ["UserPromptSubmit", "PreToolUse"] },
  fn: async (ctx) => {
    const s = load(ctx);
    const prompt = promptOf(ctx);
    if (prompt && RECOVERY_TASK.test(prompt)) {
      s.l4RecoveryTask = true;
      if (DATABASE_TASK.test(prompt)) s.l4DatabaseRecovery = true;
    }
    const cmd = commandOf(ctx);
    if (cmd && BACKUP.test(cmd)) {
      if (!s.l4DatabaseRecovery || DATABASE_BACKUP.test(cmd)) s.l4BackupMade = true;
    }
    save(ctx, s);
    return allow();
  },
});

customPolicies.add({
  name: "luna4-preserve-before-open",
  match: { events: ["PreToolUse"] },
  fn: async (ctx) => {
    const s = load(ctx);
    const cmd = commandOf(ctx);
    if (!cmd || !s.l4RecoveryTask || s.l4BackupMade || !MUTATING_OPEN.test(cmd)) return allow();
    // Keep blocking until a sufficient backup exists. Repeating this deny is
    // intentional: one warning followed by an incomplete single-file copy
    // must not silently unlock the destructive operation.
    return deny(
      `This recovery task has not preserved the original evidence yet, and this command can mutate it ` +
      `while opening or repairing it. Make a byte-preserving copy of the artifact and its sidecars ` +
      `first, perform recovery work on the copy, and use a read-only reader or hex dump on the original.`
    );
  },
});
\end{lstlisting}
\subsection{Exact replacements}
\begin{lstlisting}
// Frozen exact-replacement policy. Its narrow credential path produced a
// focused fired-policy pass after write tracking was corrected.

import { customPolicies, allow, instruct } from "failproofai";
import { load, save, commandOf, promptOf, shouldNudge, noteCommand, WRITE as GENERIC_WRITE } from "../_lib/state.mjs";

const REPLACEMENT = /\b(?:replace\b[^.\n]{0,100}\bwith\b|specified synonyms?|allowed synonyms?|placeholder values?|only edits? you may make)\b/i;
const PROTECTED = /\b(?:do not (?:modify|change|edit|touch)|leave [^.\n]{0,80} (?:unchanged|intact)|only (?:modify|change|edit)|only edits? you may make)\b/i;
const EXACT_PLACEHOLDERS = /\bplaceholder values?\b[^.\n]{0,160}\b(?:consistent|exact|same|as follows)\b/i;
const SECRET = /\b(?:api keys?|access keys?|secret keys?|github tokens?|huggingface tokens?|credentials?)\b/i;
const WRITE = /\b(?:perl\s+-p?i|sed\s+-i|python(?:3)?\b[^\n]{0,180}(?:write_text|write_bytes|open\([^\n]{0,80}['"]w))\b/i;
const DIFF = /\bgit\s+diff\b|\bcmp\b|\bsha256sum\b|\bdiff\s+(?:-[A-Za-z]+\s+)*\S+/i;
const FINAL_CHECK = /\b(?:pytest|py\.test|pdflatex|git\s+diff\s+--check|grep\s+-[A-Za-z]*q|rg\s+)[^\n]*/i;
const PATCH_WRITE = /apply_patch|\*\*\*\s+(?:Add|Update|Delete)\s+File:/;

customPolicies.add({
  name: "final-replacement-track",
  match: { events: ["UserPromptSubmit", "PreToolUse"] },
  fn: async (ctx) => {
    const s = load(ctx);
    const prompt = promptOf(ctx);
    if (prompt && REPLACEMENT.test(prompt)) {
      if (SECRET.test(prompt) && EXACT_PLACEHOLDERS.test(prompt)) s.finalSecretReplacement = true;
      else if (PROTECTED.test(prompt)) s.finalRestrictedReplacement = true;
    }
    const cmd = commandOf(ctx);
    if (cmd) {
      noteCommand(s, cmd);
      if (WRITE.test(cmd)) s.finalReplacementWrite = s.seq;
      if (DIFF.test(cmd)) s.finalReplacementDiff = s.seq;
      if (FINAL_CHECK.test(cmd)) s.finalReplacementCheck = s.seq;
    } else if (/apply_patch|\*\*\*\s+(?:Add|Update|Delete)\s+File:/.test(JSON.stringify(ctx?.toolInput || {}))) {
      s.seq = (s.seq || 0) + 1;
      s.lastWrite = s.seq;
      s.finalReplacementWrite = s.seq;
    }
    save(ctx, s);
    return allow();
  },
});

customPolicies.add({
  name: "final-replacement-before-first-write",
  match: { events: ["PreToolUse"] },
  fn: async (ctx) => {
    const s = load(ctx);
    const blob = JSON.stringify(ctx?.toolInput || {});
    const cmd = commandOf(ctx);
    const isWrite = WRITE.test(cmd) || GENERIC_WRITE.test(cmd) || PATCH_WRITE.test(blob);
    if ((!s.finalSecretReplacement && !s.finalRestrictedReplacement) || !isWrite) return allow();
    if (!shouldNudge(s, "final-replacement-before-write", 1)) return allow();
    save(ctx, s);
    return instruct(
      `Before making the first edit, treat the supplied replacement map as a byte-level allowlist. ` +
      `Change only listed source tokens to members of their own listed family; never substitute an article, ` +
      `whitespace, punctuation, quoting, or newline unless it is explicitly listed. Record each source-to-target ` +
      `pair so the final diff can be checked mechanically, and preserve every file's original EOF-newline state.`
    );
  },
});

customPolicies.add({
  name: "final-secret-replacements",
  match: { events: ["Stop"] },
  fn: async (ctx) => {
    const s = load(ctx);
    if (!s.finalSecretReplacement || !s.finalReplacementWrite || !shouldNudge(s, "final-secret-replacements", 1)) return allow();
    save(ctx, s);
    return instruct(
      `This credential-sanitization task specifies exact placeholder text. Replace each credential token ` +
      `character-for-character while preserving every surrounding byte; do not add quotes or escapes around ` +
      `a placeholder. Inspect all working-tree files, including ignored or generated files unless excluded, ` +
      `but do not rewrite history or delete refs unless explicitly requested. Verify every original value is ` +
      `absent, every placeholder is exact, and only contaminated files changed. Inspect the full patch and ` +
      `preserve each file's original end-of-file newline state.`
    );
  },
});

customPolicies.add({
  name: "final-restricted-replacements",
  match: { events: ["Stop"] },
  fn: async (ctx) => {
    const s = load(ctx);
    if (!s.finalRestrictedReplacement || !s.finalReplacementWrite) return allow();
    if (!shouldNudge(s, "final-restricted-replacements", 1)) return allow();
    save(ctx, s);
    return instruct(
      `This task supplies a closed replacement or synonym map and protects other content. Inspect the final ` +
      `diff without editing it. Mechanically parse the replacement map and compare the original and final token ` +
      `streams: every changed token must map to a member of the same explicit family, while punctuation, ` +
      `whitespace, token count, protected files, and EOF-newline state remain unchanged. Do not accept a visual ` +
      `inspection or a clean compiler log as proof. Also identify every source-to-allowed-replacement pair; ` +
      `reject invented synonyms, approximate placeholders, newline-only changes, and unrelated files. Search ` +
      `once more for every original value, then run the task's end-to-end check after the final edit. If the ` +
      `diff and check already pass, stop immediately and do not rewrite anything.`
    );
  },
});
\end{lstlisting}
\subsection{LaTeX log validation}
\begin{lstlisting}
// Frozen LaTeX log policy from the focused fired-policy pass.

import { customPolicies, allow, instruct } from "failproofai";
import { load, save, commandOf, promptOf, shouldNudge } from "../_lib/state.mjs";

const CONTRACT = /\bno\s+["'`]?overfull\s+(?:\\hbox|hbox)\b/i;
const BUILD = /\b(?:pdflatex|xelatex|lualatex|latexmk)\b/i;

customPolicies.add({
  name: "final-latex-log-track",
  match: { events: ["UserPromptSubmit", "PreToolUse"] },
  fn: async (ctx) => {
    const s = load(ctx);
    const prompt = promptOf(ctx);
    if (prompt && CONTRACT.test(prompt)) s.finalLatexContract = true;
    const cmd = commandOf(ctx);
    if (cmd && BUILD.test(cmd)) s.finalLatexBuild = true;
    save(ctx, s);
    return allow();
  },
});

customPolicies.add({
  name: "final-latex-log-validation",
  match: { events: ["Stop"] },
  fn: async (ctx) => {
    const s = load(ctx);
    if (!s.finalLatexContract || !s.finalLatexBuild || !shouldNudge(s, "final-latex-log", 1)) return allow();
    save(ctx, s);
    return instruct(
      `Inspect the log produced by the final LaTeX compile and search for the literal text ` +
      `'Overfull \\hbox' with exactly one backslash. Do not infer cleanliness from exit status. ` +
      `If it is present, make only a replacement explicitly authorized by the task, then rerun both ` +
      `the compile and literal log check. If it is absent and the allowed-replacement diff is valid, stop.`
    );
  },
});
\end{lstlisting}
\subsection{Binary equivalence}
\begin{lstlisting}
// Narrow threshold policy for reverse-engineered replacement programs. It
// avoids exposing ordinary threshold-bearing tasks to a generic optimization
// nudge.

import { customPolicies, allow, instruct } from "failproofai";
import { load, save, commandOf, promptOf, shouldNudge } from "../_lib/state.mjs";

const REIMPLEMENT = /\b(?:compiled|existing)\s+(?:an?\s+|the\s+)?(?:program|binary|executable)\b/i;
const EQUIVALENCE = /\b(?:identical|equivalent|same)\s+(?:operation|output|behavio[u]?r)\b/i;
const INDEPENDENT = /\b(?:independent|isolation|must not invoke|without (?:calling|running|using))\b/i;
const SIZE_BAR = /(?:<|under|at most|no more than)\s*\d+\s*(?:k|kb|bytes?)\b/i;
const MEASURE = /\b(?:cmp|diff|sha256sum|gzip\s*\|\s*wc|wc\s+-c)\b/i;
const CANDIDATE_BUILD = /\bgcc\b[^\n]{0,240}\b(?:mystery\.c|reversed)\b|\.\/reversed\b/i;

customPolicies.add({
  name: "final-binary-equivalence-track",
  match: { events: ["UserPromptSubmit", "PreToolUse"] },
  fn: async (ctx) => {
    const s = load(ctx);
    const prompt = promptOf(ctx);
    if (prompt && REIMPLEMENT.test(prompt) && EQUIVALENCE.test(prompt) && INDEPENDENT.test(prompt) && SIZE_BAR.test(prompt)) {
      s.finalBinaryEquivalence = true;
    }
    const cmd = commandOf(ctx);
    if (cmd && MEASURE.test(cmd)) s.finalBinaryMeasurement = true;
    save(ctx, s);
    return allow();
  },
});

customPolicies.add({
  name: "final-binary-equivalence-before-candidate-run",
  match: { events: ["PreToolUse"] },
  fn: async (ctx) => {
    const s = load(ctx);
    const cmd = commandOf(ctx);
    if (!s.finalBinaryEquivalence || !cmd || !CANDIDATE_BUILD.test(cmd)) return allow();
    if (!shouldNudge(s, "final-binary-before-run", 1)) return allow();
    save(ctx, s);
    return instruct(
      `The first candidate must be measured against the original immediately, before further approximation. ` +
      `Run both in isolated directories with the evaluator-facing build and invocation, compare every output ` +
      `artifact and stream byte-for-byte, and record the actual similarity or mismatch. If equivalence fails, ` +
      `switch to reverse-engineering the missing operation instead of polishing a visually similar renderer.`
    );
  },
});

customPolicies.add({
  name: "final-binary-equivalence",
  match: { events: ["Stop"] },
  fn: async (ctx) => {
    const s = load(ctx);
    if (!s.finalBinaryEquivalence || !shouldNudge(s, "final-binary-equivalence", 1)) return allow();
    save(ctx, s);
    return instruct(
      `Before stopping, validate the replacement program against the original over a diverse input set using ` +
      `the exact build and invocation contract. Record output differences and the compressed source size, and ` +
      `verify that the replacement remains independent when the original executable is unavailable. If any ` +
      `measured acceptance bar is missed, change the implementation method rather than declaring approximate ` +
      `behavior sufficient. Preserve the best measured artifact.`
    );
  },
});
\end{lstlisting}
\subsection{Dataset-token accounting}
\begin{lstlisting}
// Narrow successor to the broad requirement ledger. It preserves the exact
// field-reconciliation intervention that recovered the token-counting target.

import { customPolicies, allow, instruct } from "failproofai";
import { load, save, promptOf, shouldNudge } from "../_lib/state.mjs";

const DATASET = /\b(?:dataset|corpus)\b/i;
const TOKEN_COUNT = /\b(?:how many|count|number of|total)\b[^.\n]{0,100}\btokens?\b|\btokens?\b[^.\n]{0,100}\b(?:count|total|number)\b/i;
const SELECTION = /\b(?:domain|category|split|label|subset)\b/i;
const TOKENIZER = /\btokeni[sz]er\b/i;

customPolicies.add({
  name: "final-token-accounting-track",
  match: { events: ["UserPromptSubmit"] },
  fn: async (ctx) => {
    const s = load(ctx);
    const prompt = promptOf(ctx);
    if (prompt && DATASET.test(prompt) && TOKEN_COUNT.test(prompt) && SELECTION.test(prompt) && TOKENIZER.test(prompt)) {
      s.finalTokenAccounting = true;
    }
    save(ctx, s);
    return allow();
  },
});

customPolicies.add({
  name: "final-token-accounting",
  match: { events: ["Stop"] },
  fn: async (ctx) => {
    const s = load(ctx);
    if (!s.finalTokenAccounting || !shouldNudge(s, "final-token-accounting", 1)) return allow();
    save(ctx, s);
    return instruct(
      `Before reporting this dataset token count, write a compact reconciliation table. Inspect the source's ` +
      `actual distinct domain/category/split values. Use the exact requested value if present; otherwise use ` +
      `only a mapping explicitly documented by the dataset and verify the selection is non-empty. Apply the ` +
      `named tokenizer. List every text field that contributes to the requested entity, compute each field's ` +
      `subtotal, and reconcile their sum to the final integer. Do not report one field as the total.`
    );
  },
});
\end{lstlisting}
\subsection{Complete multi-answer enumeration}
\begin{lstlisting}
// Narrow successor to the generic quantifier policy. It targets tasks that
// explicitly require every valid/winning answer as separate output records.

import { customPolicies, allow, instruct } from "failproofai";
import { load, save, promptOf, shouldNudge } from "../_lib/state.mjs";

const MULTIPLE = /\b(?:multiple|more than one|all|every)\b[^.\n]{0,100}\b(?:winning|valid|correct|acceptable)\b[^.\n]{0,80}\b(?:moves?|answers?|solutions?|results?)\b/i;
const OUTPUT_EACH = /\b(?:one per line|each on (?:a|its own) line|print them all|write them all|output them all)\b/i;

customPolicies.add({
  name: "final-multi-answer-track",
  match: { events: ["UserPromptSubmit"] },
  fn: async (ctx) => {
    const s = load(ctx);
    const prompt = promptOf(ctx);
    if (prompt && MULTIPLE.test(prompt) && OUTPUT_EACH.test(prompt)) s.finalMultiAnswer = true;
    save(ctx, s);
    return allow();
  },
});

customPolicies.add({
  name: "final-multi-answer-enumeration",
  match: { events: ["Stop"] },
  fn: async (ctx) => {
    const s = load(ctx);
    if (!s.finalMultiAnswer || !shouldNudge(s, "final-multi-answer", 1)) return allow();
    save(ctx, s);
    return instruct(
      `The output contract requires every valid answer, not one witness. Enumerate the complete candidate ` +
      `set, test each candidate against the task's winning or validity condition, remove duplicates, and emit ` +
      `all surviving answers in the exact requested one-record-per-line format. Before stopping, verify that ` +
      `no additional candidate satisfies the same condition.`
    );
  },
});
\end{lstlisting}

\end{document}